# *ProVe* - Self-supervised pipeline for automated product replacement and cold-starting based on neural language models

**Andrei Ionut Damian**
andrei@lummetry.ai
Lummetry.AI

**Laurentiu Piciu**
laurentiu@lummetry.ai
Lummetry.AI

**Cosmin Mihai Marinescu**
cosmin@lummetry.ai
Lummetry.AI

## Abstract

In retail vertical industries, businesses are dealing with human limitation of quickly understanding and adapting to new purchasing behaviors. Moreover, retail businesses need to overcome the human limitation of properly managing a massive selection of products/brands/categories. These limitations lead to deficiencies from both commercial (e.g. loss of sales, decrease in customer satisfaction) and operational perspective (e.g. out-of-stock, over-stock). In this paper we propose a pipeline approach based on Natural Language Understanding, for recommending the most suitable replacements for products that are out-of-stock. Moreover, we will propose a solution for managing products that were newly introduced in a retailer's portfolio with almost no transactional history. This solution will help businesses: automatically assign the new products to the right category; recommend complementary products for cross sell from day 1; perform sales predictions even with almost no transactional history. Finally, the vector space model resulted by applying the pipeline presented in this paper is directly used as semantic information in deep learning-based demand forecasting solutions, leading to more accurate predictions. The whole research and experimentation process have been done using real life private transactional data, however the source code is available on https://github.com/Lummetry/ProVe.

## 1 Introduction and problem definition

During the past years, retailers have experienced a constant change of consumers' behavior together with a massive increase of their product portfolio. Managing the dynamics of both these situations became critical. Retailers lose sales when products are out-of-stock. Therefore, businesses would greatly benefit from an automated solution that would better predict the future demand and provide an effective product replacement recommendation (when out-of-stock; including also for newly introduced products).

Deep representation learning based on Natural Language Understanding can be successfully used to approach such challenges.

By applying it to inherent retail transactional data (e.g. checkout point-of-sales data), deep representation learning allows us to automatically – and potentially in un-supervised fashion – discover rich features of such data.

Since the introduction of the well-known word-embeddings generation algorithms in early 2015, the area of natural language processing and understanding has seen a tremendous improvement. Although it might seem that the actual state-of-the-art resides strongly on contextual embeddings and/or complex multi-head self-attention architectures, it is all based on the initial "basic" steps in the area of semantic vector space models.

Based on this rich area of research and experimentation, a lot of development has been done in the area of applying NLP/NLU approaches to commercial transaction systems, including recommender systems. One of the most common approaches, further described later in section 2, is to view a retail transaction (a list of purchased items) as a natural language object – a context window, the whole retail transactional database as the actual text corpus and the item SKU database as the "word vocabulary". This approach of creating a hypothesis similarity between the natural language representation learning and business analytics systems has proved successful in various cases and it is still a very active area of research.

The main intuition behind our proposed deep learning pipeline was that we could generate, through multiple modeling iterations, powerful-enough semantic vector space representations for each individual item (product) so that we can infer replacement (synonym) items and propose them if the original item goes out-of-stock – all of these in a self-supervised setting.

Moreover, we could use the resulted vector space representations for cold-starting of products that were newly introduced in the retailers' portfolio. This solution will help businesses: automatically assign the new products to the right category; recommend complementary products for cross sell from day 1; perform sales predictions even with almost no transactional history.

During our research, we aimed for using an efficient and scalable approach that would fully take advantage of GPU computational resources and that would easily scale for real-life projects with large transactional databases and inventories of tens of thousands of items.

## 2   Related work

We concentrated our project inception phase on:
- ✓ Reviewing semantic vector space model generation based on the most commonly known methods followed by understanding what tools and approaches we have at our disposal in order to fine-tune the embedding spaces.
- ✓ Reviewing cross-domain application of NLU in the product recommendation systems that are relying on representation learning, as well as understand their limitations.

### 2.1   Semantic vector space models

One of the most important papers relevant to this subject (beside the well-known *word2vect* (Mikolov, Chen, Corrado, & Dean, 2013)) is *"GloVe: Global Vectors for Word Representation"* (Pennington, Socher, & Manning, 2014) as it describes one of the core representation learning methods required to generate sound semantic word embeddings. The main intuition behind *GloVe* is based on the fact that statistics of word co-occurrence in a corpus is the primary source of information of the proposed unsupervised method for learning word representations. Thus, the proposed word vector generation method first constructs the matrix of co-occurrence then applies a weighted least squares regression using as target the natural logarithm of the word counts as defined by the below objective function.

$$J = \sum_{i,j=1}^{V} f(X_{ij})\,(w_i^T \widetilde{w}_j + b_i + \tilde{b}_j - logX_{ij})^2$$

(1)

Finally, as the model generates through the optimization process two sets of word vectors ($W$ and $\widetilde{W}$), the *GloVe* algorithm summarizes the two matrices in order to obtain the final semantic vector space models.

Several other aspects of interest for the construction of our hypothesis can be found in the analyzed paper and one of the most important is the correlation between the word vector dimension (actually, the model capacity) and the task-related accuracy (see Figure 3).

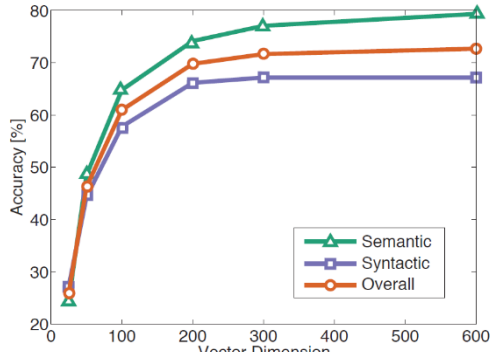

Figure 3 - Accuracy on a specific task as a function of vector size. Image from original paper by Pennington et al

## 2.2 Product recommendations

One of the early research papers that proposed the encoding of item SKUs in a similar manner to that of words is *"E-commerce in your inbox: Product recommendations at scale"* (Grbovic, et al., 2015). The authors propose a direct analogy between a certain product (item) in a basket and the focal word in skip-gram context as presented in Figure 2.

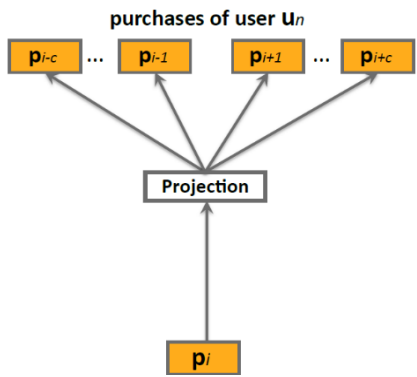

Figure 2 - The skip-gram architecture similar with original word2vec paper. Image from original paper by *Grbovic et al*

Another proposed aspect was that of enriching the vector space model with meta information such as user-identification in a similar method with *doc2vec* approach – a follow-up of the original word2vec – by Le and Mikolov (Le & Mikolov, 2014). In Figure 1 is presented this particular approach, this time being based on the bag-of-words algorithm (predicting a focal item based on the context) instead of the skip-gram (predicting the context starting from a focal item).

In terms of vector space model applications, the authors propose two different hypotheses:

- straight correlation between items cosine distances and the real-life similarities of the products (namely the **prod2vec-topK** approach)

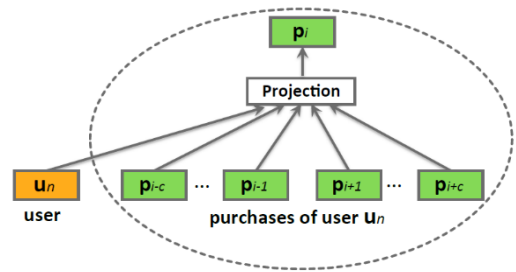

Figure 1 - Bag-of-words like approach to enrich the vector space model with user meta information. Picture based on the original paper by Grbovic et al

- a second more elaborate heuristic approach of finding viable "complementarity" baskets of products that might be sold well together. This second proposed predictive model, called **prod2vec-cluster** by the authors, leverages the hypothesis that similar items can be grouped together distance-wise. Furthermore, close clusters generated in this manner yield potentially well correlated products and picking items from different close clusters will generate a well-defined complementarity basket.

This particular work is one of the earliest and most well known in the area of applying representation learning and neural language modelling to the problem of retail recommender system.

In contrast with this work, the ACM Recommender Systems Challenge'17 was won by a more recent yet more conservative approach by Volkovs et al. In their paper "*Content-based Neighbor Models for Cold Start in Recommender Systems*" (Volkovs, Yu, & Poutanen, 2017) the authors propose a classic approach based on heavy manual featurization of the input data and full supervised approach. Although the proposed method received praises due to the high scores in the competition it is obvious that this method - relied entirely on heavily supervised dataset and hand-crafted features - **does not possess the flexibility and the scale ability potential of self-supervised systems.**

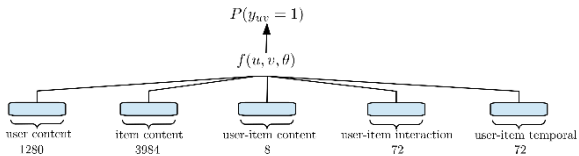

Figure 4 - Figure from paper of Volkovs et all presenting the overall architecture of their classification model using over 4000 hand-crafter features

## 2.3   Retrofitting and counter-fitting

Following the analysis of the base tool – the *GloVe* word-embeddings generation approach and its particular (and potential) cross-domain application to recommender systems - it is now required to revisit research work in the area of enriching vector space models based on retrofitting and counter-fitting. For this subject we decided to analyze three important papers: "*Retrofitting Word Vectors to Semantic Lexicons*" by Faruqui et al (Faruqui, et al., 2014), "*Counter-fitting word vectors to linguistic constraints*" (Mrkšić, et al., 2016) and the more recent work of Benjamin J. Lengerich, Andrew L. Maas and Christopher Potts "*Retrofitting Distributional Embeddings to Knowledge Graphs with Functional Relations*" (Lengerich, Maas, & Potts, 2017).

*Basic retrofitting*

In order to better understand the motivation to review the vector space models retrofitting approaches (and in a later section to review the counter-fitting approaches) we have to revisit the work done by Grbovic et *al* as well as other teams that pursued the goal of obtaining products semantic vector space models. In all above approaches the authors base their work on the *shallow* hypothesis that similar items measured by cosine distance might have a synonym-like or an entailing-like relationship. However, this shallow hypothesis not accurate in real life, neither for recommender systems based on vector space models, nor for the natural language vector space models counterparts. For example, a *GloVe-300* vector space model, pretrained on *Wikipedia* and *Gigaword* might tell us that *'adolescent'*,

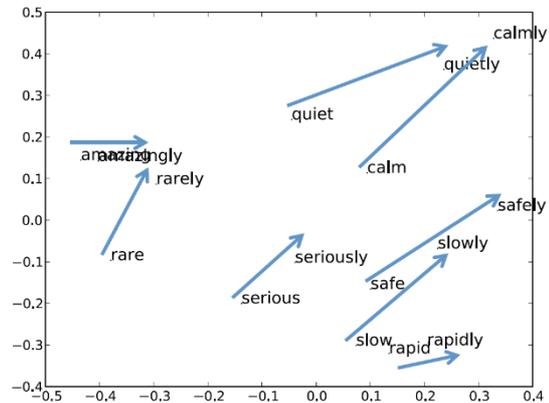

Figure 5 - Closing the distance between adjective-adverb pairs. Figure from original paper by Faruqui et al presenting the post-retrofit vector space projection

*'teenager'*, *'puberty'* are very similar to each other based on their pairwise cosine distance. However, '*adolescent'* is more similar to '*puberty'* than it is to '*teenager'*. Just using a raw embedding generated with *GloVe* or *word2vect* is not enough to get rich and reliable properties of the words and the same statement is valid also for the cross-domain applications of these methods.

This is the point where adding meta-information (similar to the *synonymity graphs* in natural language) might greatly aid in further refining the semantic power of our product vector space model. The main intuition is well defined in the work of Faruqui et al. as "*graph-based learning technique for using lexical relational resources to obtain higher quality semantic vectors*" by means of post-processing the pre-trained semantic vector space models.

$$J = \sum_{i=1}^{N} \left[ \alpha_i \|q_i - \hat{q}_i\|^2 + \sum_{i,j \in E} \beta_{ij} \|q_i - q_j\|^2 \right] \quad (2)$$

Another important aspect is that the proposed "*retrofitting*" optimization method, described by the objective function in equation 2, **is agnostic to the origins of the input vector space model**, such as original training objective or overall initial training approach. In the this equation $Q$ is the new optimized vector space model, $\hat{Q}$ is the original vector space model, and $i,j$ are pairs of words that are connected by an edge in the semantic relationship graph (knowledge graph).

The authors explain in the paper that they perform experiments using various semantic lexicons such as PPDB (Ganitkevitch, Van Durme, & Callison-Burch, 2013) and WordNet (Miller, 1995) and use these in order to improve the word vectors. Following the optimization process they evaluate the quality of the new *retrofitted* word vectors in order to determine how well they capture semantic aspects.

In another paper on the subject of counter-fitting (Mrkšić, et al., 2016) the authors take a more generalized approach by deciding to intuitively pull together the synonymous words while pushing antonymous words vectors apart. The pushing operation proposed by Mrkšić et *al* can be viewed as a similar operation to that proposed by the work of Faruqui et *al* although the optimization process is slightly different. *The counter-fitting* method proposes a three-part objective function as follows:

- the first component, described below in equation (3), of the objective function is responsible of antonym pushing by imposing a minimal *distance δ* (using a distance function *d*) for the antonymous word embeddings ($\tau$ is *max* function)

$$J_A = \sum_{u,w \in A} \tau(0, \delta - d(v'_u, v'_w)) \quad (3)$$

- the second term – equation (4) - purpose is to pull synonymous word vectors closer by reducing the distance between them to a certain margin and it is closely related with the second term from the objective function in the Faruqui et al work described by equation (2).

$$J_S = \sum_{u,w \in A} \tau(0, d(v'_u, v'_w)) \quad (4)$$

- the third and final component presented in equation (5) has the purpose of **preserving the overall vector space structure** by minimizing the difference between the original vector space word embedding pairs and the new *counter-fitted* word embeddings. This particular term is quite similar – and has the same purpose - with the first term in the objective function proposed by Faruqui et all (Faruqui, et al., 2014)

$$J_{VSP} = \sum_{i=1}^{n} \sum_{j \in N(i)} \tau(d(v'_u, v'_w) - d(v'_u, v'_w), 0) \quad (5)$$

Finally, the full objective function in equation of the *counter-fitting* approach is simply the weighted sum of the three components as follows:

$$J = w_1 * J_a + w_2 * J_s + w_3 * J_{VSP} \quad (6)$$

*Advanced retrofitting*

Recent work of Benjamin J. Lengerich, Andrew L. Maas and Christopher Potts show that some of the core assumptions of the original base *retrofitting* approach of Faruqui et al (such as that connected entities should have similar embeddings) cannot hold for any real-life applications - such as health domain knowledge graphs. In order to address the underlined limitations, the authors propose *Functional Retrofitting* - a retrofitting approach that sees pairwise entity relations as functions rather than simple embedding straight similarities. One of the first implication of this approach is that we can include both the *pull* and the *push* between embeddings based on real-life similarity or dissimilarity in a more robust and generalized way than the one presented by *Mrkšić et al*. In order to further understand the approach proposed by the authors of the *Functional Retrofitting*, let us understand the objective function in equation (7) by dissecting each of its four components. In order to simplify the explanations, we use a slightly different notation than in the original paper (Lengerich, Maas, & Potts, 2017)

$$J(Q,F) = J_{VSMP} + J_C + J_D + \sum_{r \in R} \rho_\lambda(f_r) \quad (7)$$

$$J_{VSMP} = \sum_{i \in Q} \alpha_i \|q_i - \hat{q}_i\|^2 \quad (8)$$

$$J_C = \sum_{(i,j,r) \in \varepsilon} \beta_{i,j,r} f_r(q_i, q_j) \quad (9)$$

$$J_D = \sum_{(i,j,r) \in \varepsilon^-} \beta_{i,j,r} f_r(q_i, q_j) \quad (10)$$

The first component $J_{VSMP}$ of the objective function is the semantic vector space model preservation constraint, identical to that in

$$\Psi_{\mathcal{G}}(\mathcal{Q}; \mathcal{F}) = \sum_{i=1}^{n} \alpha_i \|q_i - \hat{q}_i\|^2 + \sum_{(i,j,r) \in \mathcal{E}} \beta_{i,j,r} \|A_r q_j + b_r - q_i\|^2 - \sum_{(i,j,r) \in \mathcal{E}^-} \beta_{i,j,r} \|A_r q_j + b_r - q_i\|^2 + \lambda \sum_{r \in \mathcal{R}} \|A_r\|^2$$

Figure 6 - Linear relationship proposed by the authors of *Functional Retrofitting*. Applying *β=0* for *negative space* $\mathcal{E}^-$ and using identity as the value for the linear equation coefficient *A* gives us the objective function proposed by (Faruqui, et al., 2014) – image taken from original paper

(Faruqui, et al., 2014) and similar to equation (5) of (Mrkšić, et al., 2016). The second $J_C$ and third $J_D$ terms represent the application of relational penalty function for both the positive-relationship $\mathcal{E}$ knowledge-graph as well as for the $\mathcal{E}^-$ *negative space*.

Probably the most important difference between the work of (Faruqui, et al., 2014), (Mrkšić, et al., 2016) and the currently analyzed work of (Lengerich, Maas, & Potts, 2017) if the fact that we can use function-based relationship modelling. The authors of *Functional Retrofitting* propose in their paper two different approaches of the relation modelling – a linear relationship presented in Figure 6 and a basic fully connected neural network with one hidden layer.

Maybe one of the most important take-aways of this particular paper (Lengerich, Maas, & Potts, 2017) is exactly the cross-domain application – starting from semantic vector space models and knowledge graphs generally and finally using the *Functional Retrofitting* approach to identify potential uses for existing drugs in new diseases where those drugs were not yet employed.

## 3   First things first: The Data

The data we used in our initial experiment was based on the raw extraction of transactional and inventory information from a retail system that has both brick-and-mortar as well as online operations. The real-life transactional dataset contains more than 2M transactions of a retailer with various locations over a period of 3 years. Although the dataset contains more than 13,000 different products, we reduced the number of products to a maximum of 13,000 top sold products in order to speed-up experiment time using full in-GPU training.

Due to the size of transactional databases that often have billions of transactions each with many individual items we generate the proposed matrix of co-occurrence (MCO) employing efficient batch reading of the transactional data.

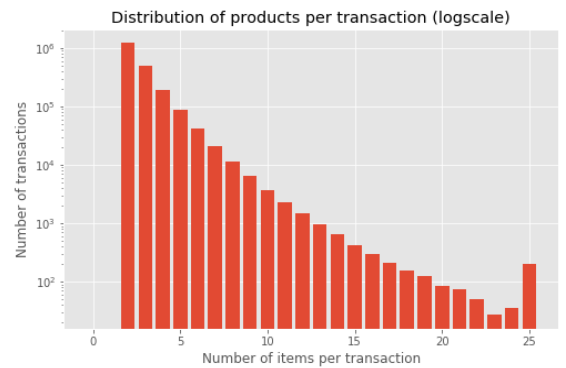

The real-life data comes within a few files generated by SQL commands and exports from an existing ERP system of our retailer. The most notable data files are the transactional database file and the metadata file. The metadata information – presented in **Error! Reference source not found.** - taken directly from a real-life production system (ERP) contains raw information minimally describing each product: the SKU (`IdemId`), the product name (`ItemName`), a unique sequential item identificatory tag (`IDE`), as well as the hierarchy information in two fields `Hierarchy1` and `Hierarchy2` that will be further used as a knowledge graph similar to WordNet (Fellbaum, 2012). There is also a secondary de-obfuscation data-source that contains for each hierarchy identification the actual name of that category. This information can also be used as a source for self-supervision in the process of creating the knowledge graph for the fine-tuning of our semantic vector space model.

The real-life transactional dataset, as previously mentioned, contains over 6M observations for over 2M different transactions. Each observation in the transactional database contains a `BasketId` identifier of the transaction as well as individual transaction detail information such as `ItemId` (and its counterpart `IDE`), a `SiteId` field that identifies the location of that particular transaction, a `TimeStamp` time-identification that is basically the same for all items of the same transaction, a quantity field, a customer identifier field and a product availability indicator. In figure … can be analyzed the distribution of products per transaction.

*Table 1 - Raw inventory information containing basic hierarchy information*

| ItemId | IDE | ItemName | H1 | H2 |
|---|---|---|---|---|
| 898234 | 2 | GREETING CARDS, A 7331335123458 | 11 | 107 |
| 891332 | 6 | GINGERBREAD HEARTS | 20 | 269 |
| 178234 | 13 | BOOK – CHRIS VOSS / NEVER SPLIT THE DIFF. | 1 | 9 |
| 565121 | 15 | BOOK – ROBERT KYOSAKI / RICH DAD POOR DAD | 1 | 0 |
| 443091 | 23 | BIO HANDBAG | 11 | 100 |
| 651651 | 24 | BOOK – ROBERT IGER / THE RIDE OF A LIFETIME | 1 | 9 |
| 290909 | 30 | BOOK - MICHELLE OBAMA / BECOMING | 1 | 9 |
| 877123 | 34 | BOOK - PAULA HAWKINS / THE GIRL ON THE TRAIN | 1 | 0 |

## 4 Overall architecture

### 4.1 Self-supervised learning for the win

As already mentioned, the winner of the ACM Recommender Systems Challenge'17 presented in section 2.2 is entirely based on supervised data and heavy manual featurization. We believe that availability of data as well as relying on well annotated data and supervised datasets will become more and more inefficient for large scale systems deployment with good generalization capacity. Worth to mention is that in the particular approach, the model devised by (Volkovs, Yu, & Poutanen, 2017) is not able to go beyond the clear and well defined purpose it was designed for and generate other insights (such as product cannibalization) out of unsupervised data.

Regarding *retrofitting* methods we already observed the objective similarity between the work of (Faruqui, et al., 2014) and the (Mrkšić, et al., 2016) where the latter improves upon Faruqui et al objectives with the addition of *antonyms push*. More interesting seems to be the parallel analysis of the previously mentioned two papers and the work of (Lengerich, Maas, & Potts, 2017). We can view equation (9) in the *Functional Retrofitting* framework as the second term in the objective function from (Faruqui, et al., 2014) and respectively the $J_S$ – equation (4) – from (Mrkšić, et al., 2016). Although equation (10) does not have an intuitive counter-part in equation (2) from (Faruqui, et al., 2014) we can consider that the *anonymity* pushing term, equation (3) from (Mrkšić, et al., 2016), is similar.

### 4.2 Two step pipeline

The two main workhorses of our pipeline are basically the initial semantic vector space generator followed by the embeddings fine-tuning model.

For the task of generating the product embeddings out of transactional information we employ either *GloVe* or *word2vec* approach to vector space generation. In our experiments section (WIP) we will argue that applying *GloVe* approach will generate better results than (Grbovic, et al., 2015) in obtaining basic semantic vector space representations of products.

While for word2vec embeddings we used the powerful and well-known *gensim* (Řehůřek & Sojka, 2010) package, for the particular *GloVe* implementation we used an extremely fast in-GPU approach that loads all co-occurrence statistics directly in the GPU computational graph and thus does not uses VRAM-RAM bus communications for batch submission, an approach forked from *Mittens* (Dingwall & Potts, 2018). Nevertheless, we have to mention that a 13,000-products inventory projected into a 128d vector space takes almost all the available 8GB VRAM of a GTX 2070 while it trains at 5 epochs per second.

Following the basic semantic vector space creation, we employ a retrofitting method using the existing meta-information in the retail databases – information such as **product categories** or even detailed category management tree-structures. This approach **leads to the decrease in the cosine distance between products that can actually replace each other in real life** in a similar manner as presented in the related work retrofitting papers address word vectors.

### 4.3 Self-supervised training and supervised testing

Although the training process is entirely self-supervised, one particular challenge that was faced was the preparation of test cases and evaluation metrics. For the development of the evaluation dataset, we have adopted a classic "supervised" approach based on manual data exploration. After manual selection of a limited list of categories based on the available meta information, a sample of products has been extracted and analyzed. For each of the target selected categories - `MUSIC, VYNIL, DVD, Board-games` - multiple products have been manually selected. Following the manual selection of product candidates for evaluation data, for each individual chosen item the top neighboring products have been prepared using cosine distance.

## 5 NLU meets BPA - semantic vector space creation

### 5.1 Semantic vector space creation

In terms of natural language models analogy to Business Predictive Analytics systems, we consider each individual basket as a "sentence" and consider each individual product id as a "word". Referring to our proposed method of constructing matrix of co-occurrence counts and applying *GloVe* approach for the vector space representation modeling we have two options:
- ✓ using a specific context window size controlled as a model hyperparameter with or without weighting between focal and each individual item;
- ✓ considering each transaction as the overall context with no distance weighting between the focal word and the context word;

Regarding the context window, it is important to note that for online shopping systems the items ordering in each individual transaction is quite important as it is directly correlated to the user journey in the website. Consequently, the particular case the definition of a hyperparameter that will control the context window size is important. For classic brick-and-mortar retailers, the order of each product in the receipt is not important as it is unlikely that it reflects the actual user sequence of actions. In our particular case of the real-life dataset we have a retailer that has over 30 locations and a website that seems to generate less than 10% of the revenue. We decided to consider each transaction as a single context for the computing of item cooccurrence counts.

With regard to the meta-data that can generate important knowledge graph information for our self-supervised approach we found out that the detail category - or the so called *"level 2 hierarchy"* - has a finer granularity than the first level hierarchical information. However, we decided to use the actual intersection of all categories as it is unknown if the hierarchy information is strictly tree-based in all retail systems.

As previously mentioned, the proposed *ProVe* model was expected to generate similar results with those of *GloVe* applied to word-vectors based on text corpuses. Just as we load a pre-trained GloVe-100 embeddings matrix with the 400,000 words vector space representations and run a neighborhood analysis for a word such as "*beatles*" and obtain "*lennon*", "*mccartney*" we obtained similar outputs from our product semantic vector space model.

The hyper-parameter selection was entirely based on the results of our exploratory data analysis on the real-life dataset combined with the know-how resulted from the natural language counterparts. For the vector space dimensionality, we have chosen 128d. We chose a maximum frequency of co-occurrences of 250, although as it can be observed from the distribution of counts, we have a maximum of over 5000 (TODO) – nevertheless we found 250 as being the best setting during our experimentation process. The number of iterations was initially on-purpose set to a large number (250,000 epochs) followed by experimentation with an early-stopping mechanism based on cosine-similarity testing of product embeddings that led to much improved results.

### 5.2 Fine tuning products semantic vectors

The second step in our pipeline is taking the initial product embeddings obtained using self-supervised optimization and further fine-tune them with self-supervised methods using meta-data obtainable from the retail systems. For the fine-tune system we applied the several methods derived from the work presented in the related work section and finally arrived at a generic cost function presented below where $J_{preserve}$ represented the vector space preservation part of the loss functions denoted by the *distance* between the initial vector representation and the modified vector space representation, $J_{relate}$ represents the objective function that pulls together the similar items based on the *membership* on the same meta-data categories denoted by the $\varepsilon_r$ graph. $J_{negate}$ is represented by an objective function that uses negative relations within $\varepsilon_n$ in order to *push* apart non-similar product vectors.

$$J = J_{preserve} + J_{relate} + J_{negate} \qquad (11)$$

$$J_{preserve} = w_p \sum_{i \in Q} D_p(q_i, \hat{q}_i) \qquad (12)$$

$$J_{relate} = w_r \sum_{i,j \in \varepsilon_r} D_r(q_i, q_j) \qquad (13)$$

$$J_{negate} = w_n \sum_{i,j \in \varepsilon_n} D_n(q_i, q_j) \qquad (14)$$

During our experimentation we experimented with various distance functions $D$ such as L1, L2 and cosine distance ($D_n = \max(m - dist(q_i, q_j), 0)$ for the case of the "push" objective function $J_{negate}$) for each of the three components of the objective function. The resulting computational graph is presented in Figure 7.

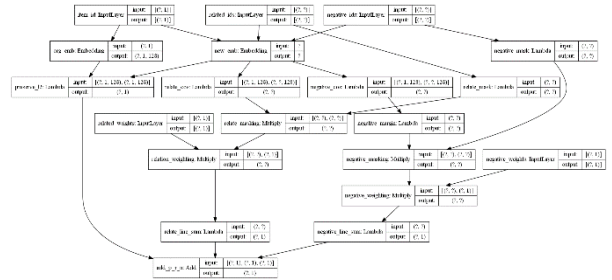

Figure 7 - Computational graph (same layout for *Tensorflow* and *PyTorch*) using cosine distance as the relation distance function

### 5.3 Hit the ground running

**For the purpose of cold-starting new items** we apply an algorithm (*Algorithm 1*) that leverages any available meta-data that we might have at the moment of the addition/creation of the new item. This information is used to create an initial vector that is then tuned accordingly in a iterative manner to a more precise position in the semantic vector space.

**Algorithm** 1 *ColdStartItem(D, N, S)*

**Input**
  *D – list of items in target category*
  *N – list of "associated items"*
  *S – list of known similar items*

**Output**
  the new item embedding vector *v*

$v \Leftarrow \frac{1}{m}\sum_{i=1}^{m} u_i$ ; $u_i \in D$
$G_N \Leftarrow$ **empty** relationship graph
$G_S \Leftarrow$ **empty** relationship graph

**if** *N* is not empty **then**
  **for each** item $w_N$ in *N*:
    add edge in $G_N \Leftarrow (v, w_N)$
  **end for**

**if** *S* is not empty **then**
  **for each** item $w_S$ in *S*:
    add edge in $G_S \Leftarrow (v, w_S)$
  **end for**

$v \Leftarrow$ retrofit with computational graph using $G_N$
$v \Leftarrow$ retrofit with computational graph using $G_S$

**Return** *v*

## 6 Evaluation and experiments

In order to apply our work to real-life experiments we decided to find the optimal tools that would greatly speed-up the particular step of retrofitting process. We used in our experiments both *Tensorflow* (Abadi, et al., 2016) as well as *PyTorch* (Paszke, et al., 2019) in order to find the optimal framework that would lead to both scalable models and fast execution. We found that *PyTorch* with its ability to execute the whole graphs in the GPU memory as well as iterate the datasets directly in GPU runs an optimization epoch in aprox. 3 seconds compared with 11 seconds *Tensorflow* in graph mode and over 20 seconds in *Tensorflow* eager execution mode. Both methods are available on the project GitHub address.

### 6.1 Metrics and overall evaluation

With regard to metrics and evaluation of this particular experiment it is important to decide how we view the model - as a classic classification or as a ranking problem. As a result, the possible metrics that can be applied to our experiment are varied and the options range from the classification metrics such as accuracy, precision, recall, F-score up to recommender systems domain-specific metrics. The initially proposed metrics were Recall@K - measures the ability of our system to recommend viable items within the first K proposed candidates - and the more important MRR@k (Mean Reciprocal Rank) - applied only for the first K candidates (if the target is ranked lower than the K then the score is 0).

$$MRR = \frac{1}{|Q|}\sum_{1}^{|Q|}\frac{1}{r_i}$$

In the above formula Q represents all the tests performed while $r_i$ represents the rank of the positive candidate (*inf* if no candidate matches the "gold" value). Finally, we decided to use only MRR as evaluation metric with various small number of candidates (K=1 and K=5), because it is mandatory to penalize the model for shallow items replacements (see again Section 2.3 – *'adolescent'*, *'teenager'* and *'puberty'* example).

### 6.2 A real example

For a better intuition, we will present an example of our pipeline on a real product ("CD / DAVID BOWIE / BLACKSTAR ").

Following the full generation of the ProVe vector space model for all product embeddings, we explored various neighbor tests in a similar manner to that of word embeddings. As previously mentioned, we selected manually each individual evaluation item in order to find those that would clearly fail at proposing a good replacement candidate just using the cosine distance neighbor search within the products vector space.

Aside from the straightforward neighborhood search we also hand-picked two other products - one that is obviously a potential replacement for the chosen product: ("*Best of David Bowie - 2002*") and another one that is clearly not a replacement ("*MASINI DIE-CAST 7.5cm III ASST*"). For these two additional products we constantly measure the distances during the various phases of our experiment.

Table 2 - The top neighbors of David Bowie's "Blackstar" based on cosine distance

| ID | DIST | NAME | H1 |
|---|---|---|---|
| 12071 | 0 | CD / DAVID BOWIE / BLACKSTAR ( | MUSIC |
| 11418 | 0.643 | CD / DAVID BOWIE / BEST OF BOW | MUSIC |
| 4416 | 0.644 | CD / COLDPLAY / 4 CD CATALOGUE | MUSIC |
| 4865 | 0.663 | CD / COLDPLAY / A HEAD FULL OF | MUSIC |
| 9745 | 0.669 | DIE-CAST CARS 7.5cm III ASST | GAMES |
| 7933 | 0.674 | CROSSWORDS / IQ FUN | BOOKS |
| 8392 | 0.676 | 13.2X2.5CM PLATE | ACCESS |
| 10135 | 0.678 | ITALIAN COOK 100 RECIPIES | BOOKS |
| 4400 | 0.681 | KISSING BOOK / IV CEL NA | BOOKS |
| 12266 | 0.697 | TIM DEDOPULOS / HERCULE POIROT | BOOKS |

In the previous example we can observe that a potential replacement for the proposed product `12071` is actually the closest product (`11418`) so this might seem a perfect fit, however we would like to have more similar products in the neighborhood of the target products and less items that clearly do not have anything in common with that product.

Following the fine tuning process we obtained results presented in Table 3 that are drastically improved versus the initial ones presented in Table 2.

Table 3 - Post retrofitting product vectors using hierarchy categories

| DIST | NAME | H1 |
|---|---|---|
| 0.000 | CD / DAVID BOWIE / BLACKSTAR ( | MUSIC |
| 0.121 | CD / RED HOT CHILI PEPPERS / T | MUSIC |
| 0.138 | CD / LANA DEL REY / LUST FOR L | MUSIC |
| 0.151 | CD / QUEEN / GREATEST HITS (20 | MUSIC |
| 0.155 | CD / COLDPLAY / A HEAD FULL OF | MUSIC |
| 0.157 | CD / QUEEN / THE PLATINUM COLL | MUSIC |
| 0.163 | CD / FLORENCE + THE MACHINE / | MUSIC |
| 0.164 | CD / QUEEN / BOHEMIAN RHAPSODY | MUSIC |
| 0.165 | CD / LED ZEPPELIN / REMASTERS | MUSIC |
| 0.165 | CD / PINK FLOYD / THE WALL (Re | MUSIC |

## 7 Conclusions and further work

Retailers can greatly benefit from our research as it gives them a better understanding of their product portfolio together with concrete insights about how to: increase sales (by offering suitable product replacements to out-of-stock items), to better manage product categories (based on actual buying behavior of products and the fine tuning of products semantic vector space), better manage newly introduced products (cold starting).

Moreover, the vector space model resulted by applying the pipeline presented in this paper is directly used as semantic information in deep learning-based demand forecasting solutions, leading to more accurate predictions.

All our research and work are centered around an efficient and scalable approach that would fully take advantage of GPU

computational resources. In order for our research to be fully adopted by real-life business projects our focus is on scalability especially when dealing with large transactional databases and inventories of tens of thousands of items.

Further experimentation work is required at this point to tune and test the models as well as finalize production-grade version of the proposed `get_item_replacement` as well as the `cold_start_item` method. More research and experimentation are also required in the area of determining an optimal heuristic method for automatically choosing the re-weighing factors of the objective function components.

## Acknowledgements

This work was supported by grant no 9/221_ap2/23.12.2019/SMIS-129090 , under the financing programme Competitiveness Operational Program (COP) Action 2.2.1 - "Supporting the growth of added value generated by the ICT sector and innovation in the field through the development of clusters" co-financed by the European Regional Development Fund.